\documentclass[conference]{IEEEtran}

\IEEEoverridecommandlockouts

\usepackage[left=0.75in,
            right=0.75in,
            top=1in,
            bottom=1in]{geometry}

\usepackage{cite}
\usepackage{amsmath,amssymb,amsfonts}
\usepackage{algorithmic}
\usepackage{graphicx}
\usepackage{textcomp}
\usepackage{xcolor}
\usepackage{multirow}
\def\BibTeX{{\rm B\kern-.05em{\sc i\kern-.025em b}\kern-.08em
    T\kern-.1667em\lower.7ex\hbox{E}\kern-.125emX}}
\begin{document}

\title{How Do Pedophiles Tweet? Investigating the Writing Styles and Online Personas of Child Cybersex Traffickers in the Philippines}

\author{\IEEEauthorblockN{Joseph Marvin Imperial}
\IEEEauthorblockA{\textit{National University} \\
Manila, Philippines \\
jrimperial@national-u.edu.ph}
}

\maketitle

\begin{abstract}
One of the most important humanitarian responsibility of every individual is to protect the future of our children. This entails not only protection of physical welfare but also from ill events that can potentially affect the mental well-being of a child such as sexual coercion and abuse which, in worst case scenarios, can result to lifelong trauma. In this study, we perform a preliminary investigation of how child sex peddlers spread illegal pornographic content and target minors for sexual activities on Twitter in the Philippines using Natural Language Processing techniques. Results of our studies show frequently used and co-occurring words that traffickers use to spread content as well as four main roles played by these entities that contribute to the proliferation of the child pornography in the country. 

\end{abstract}

\begin{IEEEkeywords}
pornography, tweets, cybersex, persona, natural language processing
\end{IEEEkeywords}

\section{Introduction}

\begin{quote}
    \textit{"Child abuse casts a shadow the length of a lifetime."}
    
    - Herbert Wood
\end{quote}

Cybersex or computer sex is an activity where two or more people, anonymous in some cases, connect over the Internet to engage sexually gratifying performances \cite{miller2013principles}. Activities such as sharing, watching, downloading and trading explicit online content across websites and social media platforms such as Facebook, Twitter, and Instagram are all under the umbrella term of cybersex \cite{cooper1998sexuality}. Behaviors exhibited in cybersex activities include solitary acts of self pleasure, consensual interactions, and to coercive and forceful activities which often considered as rape \cite{southern2008treatment}. 

In its essence, cybersex allows exploration of sexual urges and private fantasies while maintaining anonymity \cite{young2008internet} as well as providing a safe space for physically-separated partners to connect over the Web and continue to be sexually intimate \cite{miller2013principles}. However, in a moral and ethical point-of-view, the conduct of cybersex activities should only be between consenting and legal-aged partners. Non-consensual cybersex often target extremely underprivileged women and minors where the produced media are peddled, trafficked, and sold worldwide. Although most justice and intelligence agencies in countries around the world enforce strict laws on minors involved in cybersex activities, the problem still pose as one the major challenges for poor, developing areas in Southeast Asia, Africa, and South America where they are often labelled as \textbf{hotspots} of child sex tourism from 2014 to 2016 \cite{asia_times_2020}.

\subsection{Proliferation of Child Pornography in Twitter}
There are multiple environments where cybersex, both consensual and non-consensual, are often mediated and spread. Internet chat rooms and instant messaging applications are common grounds for these activities. However, in the recent years, social media platforms such as Twitter have been used more and more by illegal cybersex peddlers and traffickers since it offers anonymity under the guise of fake accounts \cite{guilbert_2018,bevan_2020}. In addition, Twitter allows these accounts to share images and videos seamlessly as well as having the option to \textit{privatize} accounts. \textbf{Pedophiles}, or a group of people who are sexually attracted to children, use these features to maintain a close circle of similar-minded individuals and to stay hidden from public eye. Although Twitter follows strict policies\footnote{Twitter Policies: www.help.twitter.com/en/rules-and-policies/twitter-rules} in maintaining a safe environment by banning users for any type of abuse, child sexual exploitation, and sexual assault, accounts of pedophiles and illegal cybersex peddlers still surge in number \cite{solon_2020}. 

In the Philippines, the Cybercrime Prevention Law was signed in 2012 which aims to reduce computer-related crimes including child pornography and other illegal cybersex activities. However, in the recent years, the Cybercrime Law did little to nothing to alleviate proliferation of child pornography as country topped the latest survey by United Nations Children's Fund on global sources of child sex abuse materials in 2018 \cite{unicef2018situation}. According to the report, the proportion of internet addresses hosting child pornographic materials in the Philippines tripled in scale starting from 2017. Twitter has become one of the most used platform in the Philippines that serves as a breeding ground and medium of pedophiles to spread child pornographic content. These individuals hide their identity using multiple fake accounts colloquially known as \textbf{alter} or alternate accounts. In the same manner, the term \textbf{Alter Twitter} has become popularly known in the country as a Twitter community of Filipino individuals using anonymous accounts to conduct, share, and exploit sexual content and activities \cite{piamonte2020virtual}. 


In recognizing the need for further research efforts in mitigating the spread of child pornographic content, this paper investigates the general writing styles of pedophiles and cybersex traffickers in Twitter, and the roles that they often conform using the platform. We perform natural language processing techniques over a dataset composed of a year's worth of child pornographic tweets collected from the Twitter accounts of pimps, peddlers, and traffickers in the Philippines.

\begin{figure}[!htbp]
    \centering
    \includegraphics[width=0.45\textwidth]{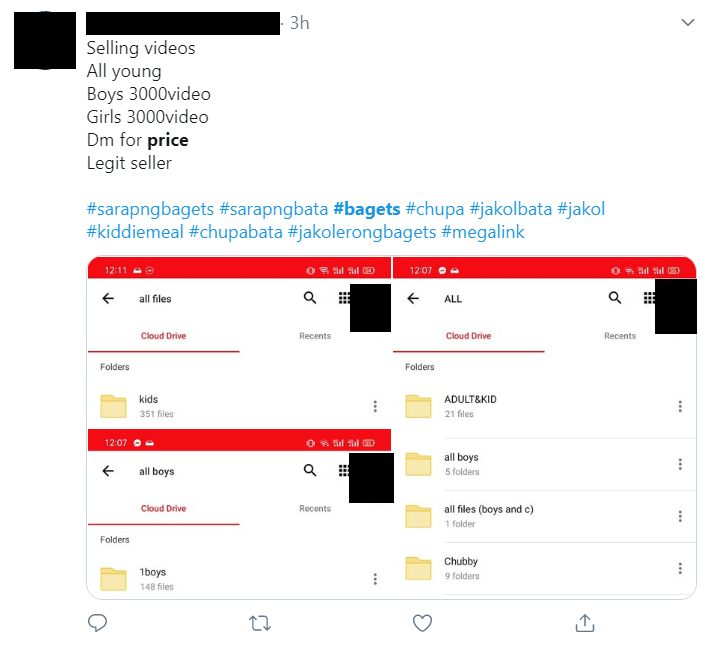}
    \caption{Example tweet showing a peddler marketing child pornographic material on Twitter.}
    \label{fig:screencap}
\end{figure}

\section{Related Works}

\subsection{Writing Styles in Twitter}
The challenge of analyzing writing styles such as authorship attribution in social media platforms is one of the most interesting tasks in natural language processing. \cite{ray2014style} defines writing style as a grammatical choice that writers make which adheres to \textbf{norms} and \textbf{social identity}. An individual's writing styles is composed of choice of select words, sentence and paragraph structure, and symbols that are used to convey the a message effectively \cite{sebranick2006writers}.

Existing writing styles in the web vary by a large scale since users are free to express themselves and there are no formal rules to follow. In addition, other elements of writing in social media platforms such as the use of emoticons to adds complexity to the task \cite{maeda2012twitter}. The use of social media platforms like Twitter allows researchers in various fields perform deeper analysis on factors that can affect writing such as gender \cite{burger2011discriminating}, user personality \cite{schwartz2013personality}, and mental illness \cite{de2013predicting}. Inclusion of these factors paved way for more research efforts in understanding negative social media interactions such as forms such abuse like racism and sexism \cite{clarke2017dimensions} and bullying \cite{lee-etal-2018-comparative}.

\subsection{Themes in Twitter}
Works on identifying salient and underlying themes conveyed in large volumes of social media data have also intrigued researchers on the field. In contrast to writing styles which focus on how each tweet is constructed using elements such as hashtags, emoticons, and use of symbols, thematic analysis captures the \textbf{representations of the texts} by uncovering \textbf{topics} commonly extracted using unsupervised machine learning algorithms to generate topic models \cite{blei2003latent,lau2012line}. These topic models allow us to have an overview of important topics (or themes) and supporting topic words present in the document \cite{blei2012probabilistic}. In the Philippine local setting, the works of \cite{ligutom2016using}, \cite{gorro2017qualitative}, and \cite{maceda2017corpus} all focused on the use of topic models to extract themes present from collected typhoon and earthquake-related Twitter data which can be used to improve the disaster risk reduction landscape and response of the country.


\section{Child Pornography-Related Tweets}
For this study, we collected over 69,675 raw tweets related to child sex trafficking and peddling in Twitter from October 2019 to July 2020, over a year's worth of data. We used a bounding-box feature from the Twitter API to capture tweets only published within the area of the Philippines. In addition, we used hashtags such as \textit{\#bagets} (colloquial term for the word 'children') and \textit{\#sarapngbagets} (conveys sexual desire for children) which were reported to be commonly used by child sex traffickers as bookmarks or subject tags for their tweets \cite{gavilan_2020}. After cleaning and removal of retweets and duplication, only 32,899 unique tweets were left for the analysis proper.

\section{Writing Style Analysis}
For the writing style analysis, we conduct two methods: the \textbf{word cloud visualization} for getting a bird's eye view of the most frequent words present within the collected data and mapping of \textbf{trigram co-occurrence network} for understanding series of word connections used to spread child pornographic content.

\begin{figure}[!htbp]
    \centering
    \includegraphics[width=0.40\textwidth]{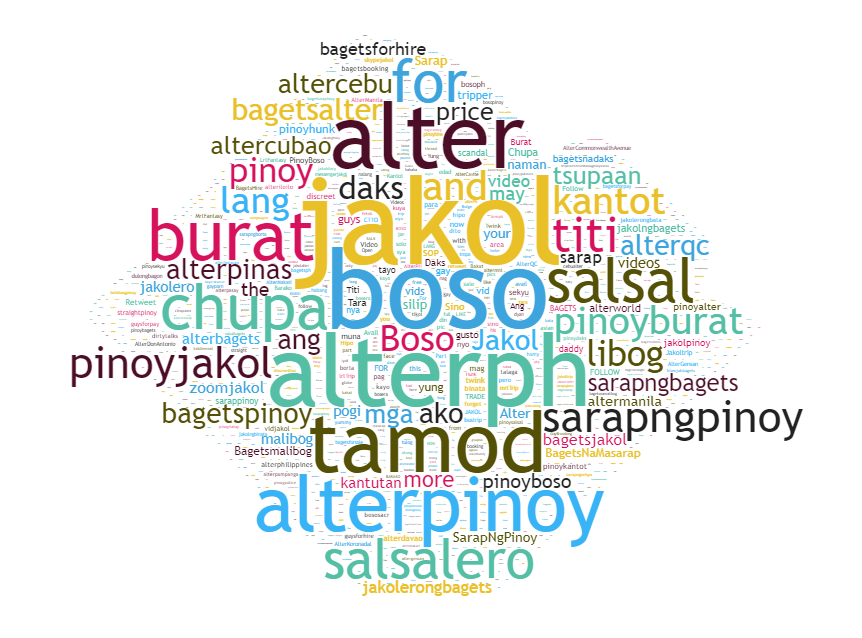}
    \caption{Word cloud visualization of top-occurring / high-frequency words from the data.}
    \label{fig:wordcloud}
\end{figure}

The word cloud visualization in Figure~\ref{fig:wordcloud} showcases top used terminologies in tweets with respect to the size of each word. The word \textit{jakol} or \textit{masturbation}, \textit{tamod} or \textit{semen}, and \textit{boso} or \textit{voyeur} are seemingly three of the most used words in the context of child pornography. In addition, the hashtag \textit{\#alterph} is also often appended in tweets to signal that the account used for uploading content is an \textit{alter} account with the suffix \textit{ph} indicating the user is in the Philippines and prefers interaction with users also coming from the same country. Action words are frequently used in context such as \textit{chupa} or \textit{fellatio} and \textit{salsal} or \textit{motion of stimulating a man's penis} as well as words used for targeting children such as \textit{bagets for hire} or \textit{children for hire} and \textit{altergc} which means \textit{alter groupchat}, indicating that there are also other platforms where videos and contents are shared and not just in Twitter.

\begin{figure}[!htbp]
    \centering
    \includegraphics[width=0.40\textwidth]{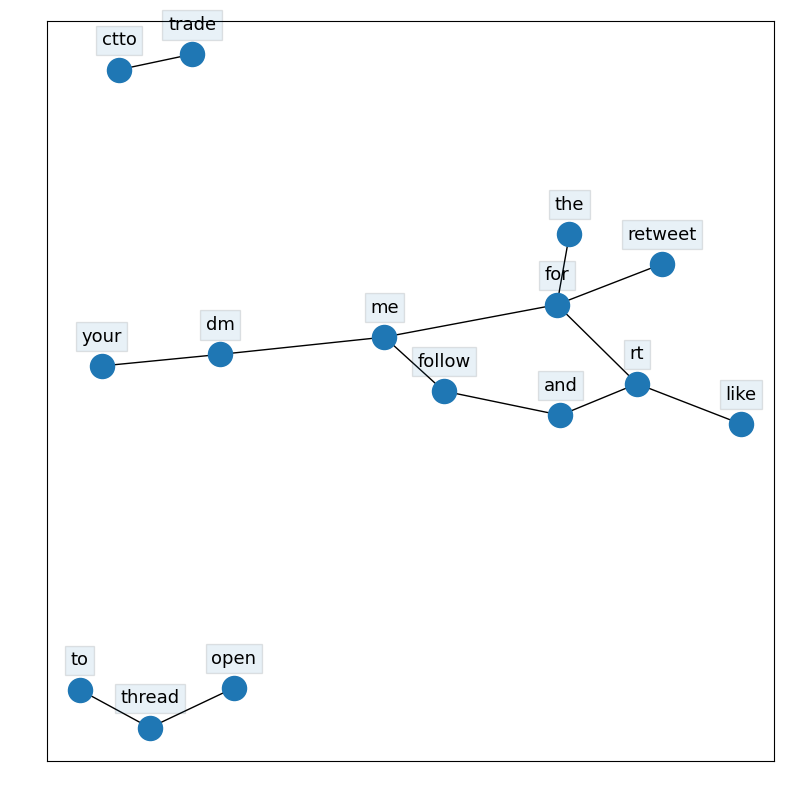}
    \caption{Trigram network indicating terms that frequently co-occur in child pornographic tweets.}
    \label{fig:network}
\end{figure}

Figure~\ref{fig:network} describes the chain reaction-like structure of words that are co-occurring or are seen together in semantically similar tweets. From a corpus containing 2,498 unique words, only three subgraphs are formed which signify that the overall lexicon used by pedophiles and child traffickers are somewhat limited in a way that terms are often reused repetitively. From the figure, the first subgraph on the upper left contains only two connected words, \textit{ctto} or \textit{credits to the owner} and \textit{trade}. These two words describe user accounts that share tweets by giving unofficial crediting of the source of contents as well as the notion of exchanging resources by trading links of online repositories where videos are stored as seen in Figure~\ref{fig:screencap}. The largest subgraph in the middle, on the other hand, contains terminologies forming sequences denoting instruction for proliferation and attention such as \textit{follow and rt (retweet)}, \textit{dm me (message me)}, and \textit{follow me}. And lastly, the third subgraph on the lower left with word sequence \textit{open thread to} denotes the spread of content to be in the form of threads or series of posts. Overall, these subgraphs model how posts containing child pornographic content such as lewd photos and videos are structured. It also describes how users behind alter accounts sway other users to spread their malicious content by convincing them to use Twitter's interaction features such as (a) retweets for sharing and (b) likes for increasing the exposure of the content to a wider audience.


\begin{table}[!htbp]
  \caption{Roles of Sexual Predators Based on Thematic Word Usage.}
  \label{tab:themes}
  \begin{tabular}{|l|l|}
    \hline
    \bf Persona &\bf Vocabulary\\
    \hline
    
    \multirow{2}{*}{The Propagator} & rt, follow, vid, retweet, videos, like, link, jakol, \\ 
    &  post, comment\\
    \hline
    
    \multirow{2}{*}{The Peddler} & bagets, dm, price, pic, area, php, avail, looking, \\
    & pls, willing\\ 
    \hline
    
    \multirow{2}{*}{The Social} & tara, face, pogi, pm, jakol, jan, pwede, want, sino \\ 
    & tayo\\ \hline
    
    The Voyeur & boso, face, tara, pic, dm, bagets, baby, sarap, cr, boys\\ 
    
  \hline
\end{tabular}
\end{table}

\section{Persona Analysis}
Aside from just analyzing the overall stylistic writing patterns of potential pedophiles and child sex traffickers on Twitter, we want to understand deeper roles played by these entities in the platform. To do this, we trained a short-text clustering model using the Gibbs Sampling Dirichlet Multinomial Mixture (GSDMM) \cite{yin2014dirichlet} trained from the pre-processed tweet corpus. The GSDMM model aggregates words into clusters or groups that are similar to each other in terms of usage and meaning. As seen in Table~\ref{tab:themes}, we obtained four main homogeneous clusters symbolizing four different online personas or roles played by users behind alter accounts that are tied with child pornography. 

From the table, each persona has its own unique set of thematic words forming an underlying vocabulary used for specific purposes. First, we have the \textbf{Propagator} which is mainly responsible for spreading child pornographic content in the platform. Keywords often used by this type of user are similar to the ones highlighted in Figure~\ref{fig:network} such as \textit{rt or retweet}, \textit{follow}, \textit{like}, \textit{post}, and \textit{comment}. Next, the \textbf{Peddler} which is responsible for the hidden market or trading, buying, and selling of child pornographic content. Keywords often used by peddlers are \textit{dm or direct message}, \textit{price}, \textit{php}, \textit{avail} for their business transactions and \textit{looking} and \textit{willing} for enticing possible victims who are willing to trade sexual content such as photos and videos for money. Third, the \textbf{Social} acts as someone who encourages users to meet physically or digitally for activities such as \textit{jakol} or \textit{masturbation}. This persona often uses descriptive words such as \textit{pogi} or \textit{handsome} as well as semi-coercive words such as \textit{tara} or \textit{let's go} and \textit{sino pwede jan?} or \textit{who is available?} to convince potential users having the same interests. Lastly, we have the \textbf{Voyeur} which often targets minors for their voyeuristic content. The main keyword used by this persona is \textit{boso} or the act of spying undressed or naked people for sexual pleasure and often used in tweets with targets such as \textit{bagets} or \textit{children}, \textit{baby}, and \textit{boys}. This persona also frequently makes use of the word \textit{cr} or \textit{comfort room} where hidden camera are often installed.

\section{Ethical Considerations}
This study makes use of extremely sensitive data involving sexual words that are often used to target minors. However, the proponents felt compelled to do this type of study as something has to be done in order to understand and be able to alleviate the problem of child pornography landscape in the Philippines. In addition, for the safety of minors, no personal information is revealed in any part of this document.

\section{Conclusion}
In the Philippines, child pornography and other illegal cyber-sex activities are widespread especially on social platforms like Twitter where users can hide behind anonymous accounts. In order to further gain understanding and deeper insights for the reason behind the rapid proliferation of child pornographic content online, we used three types of analysis, namely word cloud visualization, trigram co-occurrence analysis, and persona analysis. Results show basic terminologies often used by child traffickers and peddlers that often co-occur with each other. In addition, these entities can be classified into four possible roles or online personas based on their vocabulary use. Continuation of this study involve partnership with local government units concerned with cybercrime prevention and child protection to track down active child pornography peddlers and traffickers.

\bibliographystyle{ieeetr}
\bibliography{references}

\end{document}